\documentclass[sigplan,nonacm]{acmart}

\usepackage[utf8]{inputenc}
\usepackage[T1]{fontenc}
\usepackage{mathptmx}
\usepackage{etoolbox}
\usepackage{url}
\usepackage{xcolor}
\usepackage{booktabs}
\usepackage{bm}
\usepackage{hyperref}
\usepackage{algorithm}
\usepackage{algpseudocode}
\usepackage{multirow}
\usepackage{makecell}
\usepackage{graphicx}
\usepackage{dcolumn}
\usepackage{bm}
\usepackage{enumitem}
\setlist{nolistsep}
\usepackage[capitalise]{cleveref}

\usepackage{hyperref}

\AtBeginDocument{%
  }

\setcopyright{acmlicensed}
\copyrightyear{2018}
\acmYear{2018}
\acmDOI{XXXXXXX.XXXXXXX}
\acmConference[BIOKDD 2025]{24th International Workshop on Data Mining in Bioinformatics}{June 03--05,
  2018}{Woodstock, NY}
\acmISBN{978-1-4503-XXXX-X/2018/06}

\begin{document}

\title{Advances in Protein Representation Learning: Methods, Applications, and Future Directions}

\author{Viet Thanh Duy Nguyen}
\affiliation{
  \institution{The University of Alabama at Birmingham}
  \city{Birmingham}
  \state{Alabama}
  \country{USA}
}
\email{dvnguye2@uab.edu}

\author{Truong-Son Hy$^*$}
\affiliation{
  \institution{The University of Alabama at Birmingham}
  \city{Birmingham}
  \state{Alabama}
  \country{USA}
}
\email{thy@uab.edu} \thanks{$^*$: Corresponding Author}

\renewcommand{\shortauthors}{Nguyen and Hy}

\begin{abstract}
Proteins are complex biomolecules that play a central role in various biological processes, making them critical targets for breakthroughs in molecular biology, medical research, and drug discovery. Deciphering their intricate, hierarchical structures, and diverse functions is essential for advancing our understanding of life at the molecular level. Protein Representation Learning (PRL) has emerged as a transformative approach, enabling the extraction of meaningful computational representations from protein data to address these challenges. In this paper, we provide a comprehensive review of PRL research, categorizing methodologies into five key areas: feature-based, sequence-based, structure-based, multimodal, and complex-based approaches. To support researchers in this rapidly evolving field, we introduce widely used databases for protein sequences, structures, and functions, which serve as essential resources for model development and evaluation. We also explore the diverse applications of these approaches in multiple domains, demonstrating their broad impact. Finally, we discuss pressing technical challenges and outline future directions to advance PRL, offering insights to inspire continued innovation in this foundational field.
\end{abstract}


\keywords{Protein Representation Learning, Protein Language Models, 3D Protein Representations, Multimodal Protein Representations, Protein Complex Representation, Protein Databases, Protein Property Prediction, Protein Structure Prediction, Protein Design and Optimization, Drug Discovery}

\maketitle

\section{Introduction}
\label{sec:introduction}

\begin{figure}[t]
\centering
\includegraphics[width=\columnwidth]{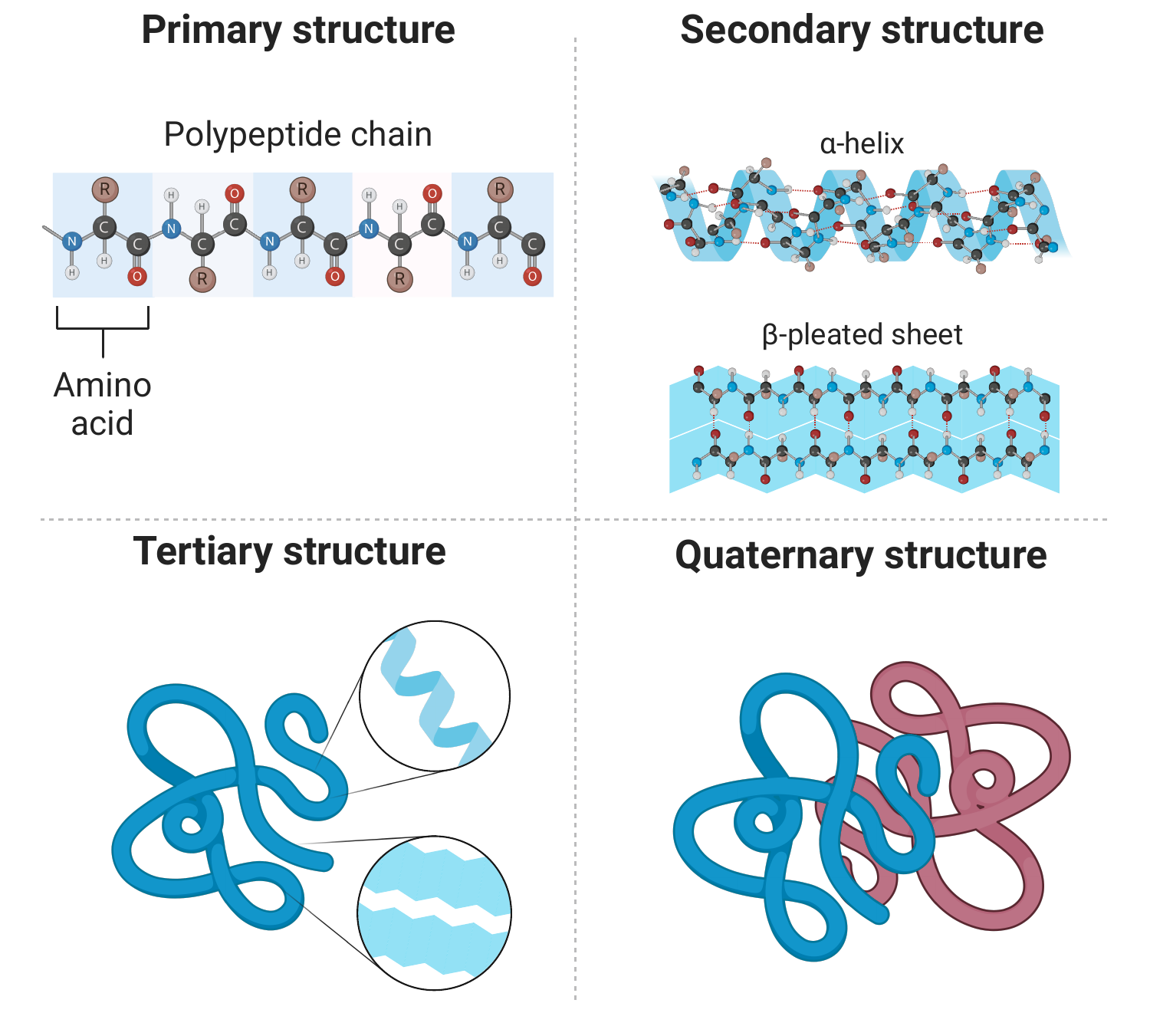}
\vspace{-0.25cm}
\caption{\label{fig:protein_levels} The four levels of protein structure, organized by increasing complexity within the polypeptide chain. Primary structure refers to the specific sequence of amino acids. Secondary structure involves local folding patterns, such as $\alpha$-helices and $\beta$-sheets. Tertiary structure represents the overall three-dimensional conformation of a single polypeptide chain. Quaternary structure describes the assembly and interactions of multiple polypeptide chains within a protein complex.}
\end{figure}

Representation learning offers a powerful framework for converting high-dimensional, complex data into compact and informative embeddings that retain essential patterns while enabling efficient computation. In the context of protein science, where sequences and structures exhibit immense diversity and complexity, such representations are crucial for supporting both predictive and generative tasks. Proteins are essential biomolecules that drive nearly all cellular processes, from catalyzing biochemical reactions to providing structural integrity and mediating signal transduction. A deep understanding of proteins is critical for numerous scientific and medical applications, including drug discovery, disease mechanism elucidation, and the design of novel enzymes or therapeutics. Central to this understanding is the recognition that the function of a protein is intimately related to its structure, which is organized hierarchically into four levels: primary (amino acid sequence), secondary (local folding motifs), tertiary (three-dimensional conformation) and quaternary (multi-subunit assembly), as illustrated in \cref{fig:protein_levels}.

Despite decades of research, decoding the structure–function relationship remains a profound scientific challenge. The mapping from sequence to structure is highly nonlinear and context-dependent, with even minor sequence variations potentially leading to significant changes in folding and function. Furthermore, proteins operate in dynamic, heterogeneous cellular environments, which adds additional complexity to their accurate modeling. The vast combinatorial space of possible sequences and conformations renders exhaustive experimental characterization infeasible and stretches the limits of traditional computational techniques. In response, Protein Representation Learning (PRL) has emerged as a transformative approach. By encoding protein sequences, structures, and functions into expressive fixed-length vectors, PRL enables scalable and data-driven analysis. Leveraging advances in deep learning and large-scale protein datasets, PRL models have achieved state-of-the-art results in structure prediction, function annotation, and de novo protein design. These developments are reshaping our ability to model, interpret, and engineer proteins with unprecedented precision and scale.

\begin{figure*}[ht]
\centering
\includegraphics[width=0.7\textwidth]{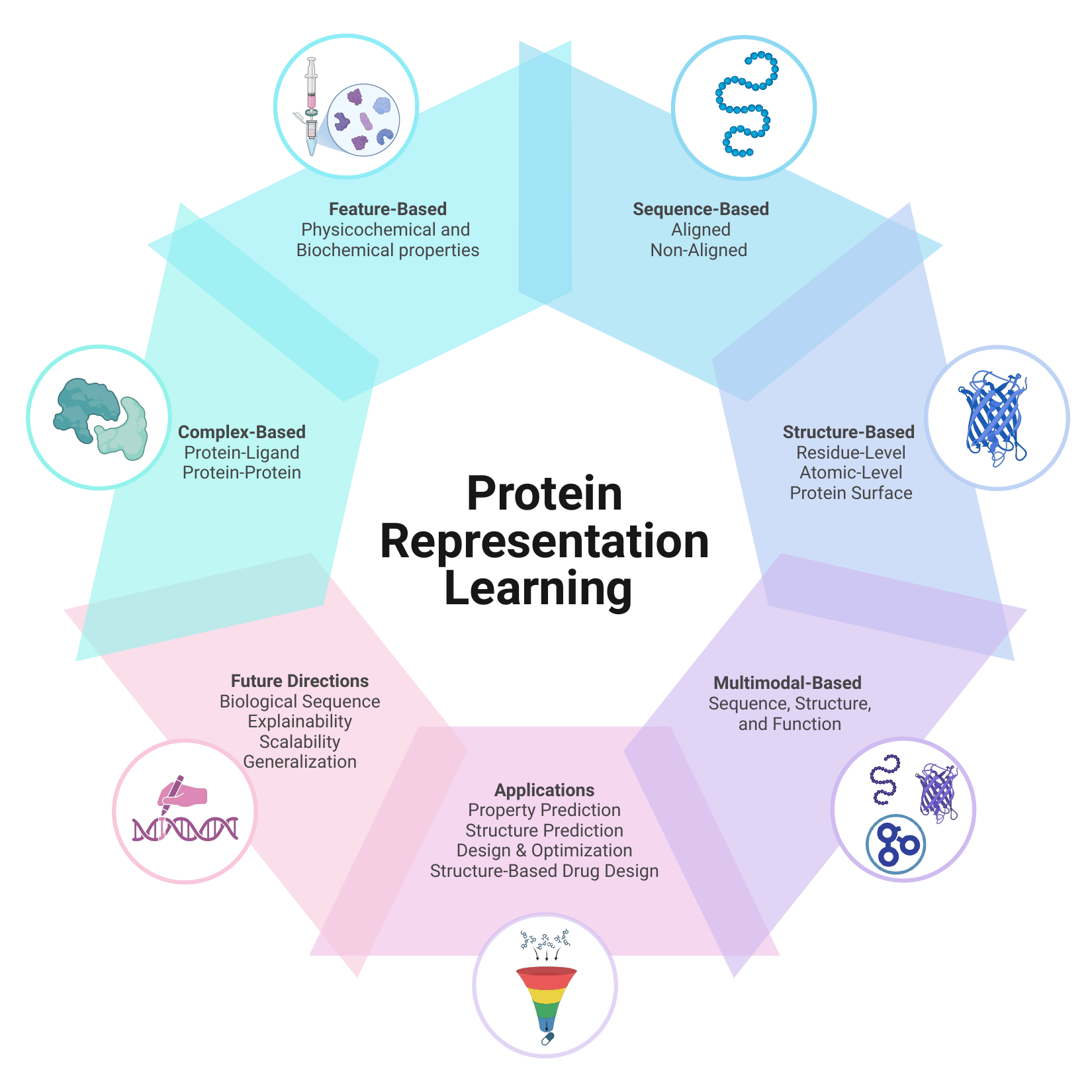}
\vspace{-0.25cm}
\caption{\label{fig:overview} An overview of the key components and themes discussed in this review. The figure highlights the interconnections and relationships between the main topics, providing a comprehensive visual summary of the review's scope.}
\end{figure*}

Although numerous surveys have been conducted in the field of PRL, most have been developed from the perspective of biological applications, model architectures, or pretext tasks \cite{wu2022survey, xiao2025protein, heinzinger2025teaching}. Although these reviews provide valuable information, they often focus on specialized aspects of PRL, making them less accessible and applicable to researchers outside of AI and computer science. In contrast, our review categorizes PRL research based on the specific modalities utilized, such as sequence, structure, and function, offering a structured framework for understanding how these data types are applied in different approaches. As illustrated in \cref{fig:overview}, this organization provides a clear pathway for researchers in various fields to identify PRL methods that align with their data types and research needs. By emphasizing the role of each modality and its relevance in diverse domains, this review bridges the gap between computational advancements and real-world applications, serving as a valuable resource for interdisciplinary researchers seeking to leverage PRL effectively.

In general, the structure of this paper is organized as follows: it begins with Section \nameref{sec:introduction}, which provides context on PRL and its importance in understanding the structure and function of the protein. Section \nameref{sec:feature} explores the early methods that encode proteins on the basis of their physical properties. Section \nameref{sec:sequence} discusses both non-aligned and aligned sequence methods for capturing protein sequence information. Section \nameref{sec:structure} delves into residue-level, atomic-level protein surface representations, along with learning of symmetry-preserving and equivariant representations, highlighting the critical role of structural data in PRL. Section \nameref{sec:multimodal} examines frameworks that integrate sequence, structure, and functional data to generate enriched representations. Section \nameref{sec:complex} covers the representations of protein-ligand and protein-protein complexes, focusing on the modeling of molecular interactions. To facilitate comparison, we provide a structured overview of the strengths and limitations of each approach in \cref{tab:prl_summary}, offering insights to guide the selection of appropriate PRL strategies based on specific biological and computational requirements. Section \nameref{sec:databases} introduces key resources for sequences, structures, and functions that support PRL. Section \nameref{sec:applications} highlights practical implementations of PRL, including protein property prediction, protein structure prediction, protein design and optimization (e.g., ligand-binding proteins, enzymes and antibodies), and structure-based drug design. Section \nameref{sec:future} discusses key challenges such as expanding PRL to DNA / RNA representation learning, improving the scalability and generalization of the model, and improving the explainability, with the aim of inspiring future research and advances in the field. Finally, Section \nameref{sec:conclusion} provides a summary of the key insights from this review and reflects on the broader impact of PRL, emphasizing its potential to advance protein science and biomedical applications.

\section{Feature-Based Approaches}
\label{sec:feature}

Feature-based protein representations encode sequences in structured numerical vectors by leveraging predefined biochemical, structural, or statistical properties of amino acids. These hand-crafted descriptors have historically played a fundamental role in computational biology, enabling a wide range of machine learning applications, including protein classification, subcellular localization, and prediction of molecular interactions. To facilitate the systematic extraction and application of these descriptors, researchers have developed software frameworks such as iFeature \cite{chen2018ifeature} and PyBioMed \cite{dong2018pybiomed}. These toolkits offer unified implementations of diverse feature encoding schemes, along with integrated support for feature selection, clustering, and dimensionality reduction. Widely adopted descriptors, grouped by the types of information they capture, include:

\begin{itemize}[noitemsep,topsep=0pt]
    \item \textbf{Composition-Based Descriptors}: These encodings quantify the overall amino acid content of a sequence without accounting for residue order. Representative examples include \textit{Amino Acid Composition (AAC)}, which computes the normalized frequency of each of the 20 standard amino acids, offering a simple but informative global feature vector. \textit{Dipeptide Composition (DPC)} extends this idea by measuring the frequencies of all contiguous amino acid pairs (400 possible combinations), capturing short-range sequence patterns that may reflect local structural motifs or functional sites.

    \item \textbf{Sequence-Order Descriptors}: These descriptors integrate sequential dependencies between residues to better capture biological relevance. A commonly used example is \textit{Pseudo-Amino Acid Composition (PseAAC)} \cite{shen2008pseaac}, which enhances traditional AAC by including correlation factors based on physicochemical properties between residues in various lags. This allows the model to retain information on sequence order and residue interactions, making it particularly effective for tasks such as subcellular localization and functional classification.

    \item \textbf{Evolutionary Descriptors}: These features encode residue conservation patterns by aligning a query sequence with homologous sequences in a database. The resulting \textit{Position-Specific Scoring Matrix (PSSM)} \cite{altschul1997gapped} represents the likelihood of observing each amino acid at every position, capturing functionally important and evolutionarily conserved regions. PSSM-derived features are widely used in structure prediction, active site identification, and protein family classification.

    \item \textbf{Physicochemical Descriptors}: These are based on experimentally measured or computationally derived amino acid properties to capture biochemical characteristics of the sequence. \textit{AAindex} \cite{kawashima2000aaindex} descriptors use numerical scales, such as hydrophobicity, polarity, isoelectric point and steric parameters, to represent individual residues. These values are then aggregated across the sequence using statistical measures (e.g., mean, standard deviation, autocorrelation). \textit{Z-scales} \cite{sandberg1998new} offer a compact alternative by summarizing multiple physicochemical indices into five orthogonal dimensions using principal component analysis, providing a low-dimensional yet expressive representation.
\end{itemize}

Although feature-based methods have played an important role in early protein modeling, they have notable limitations \cite{yang2018learned}. One of the primary challenges lies in selecting the appropriate physicochemical properties to represent amino acids, given the vast number of available descriptors and the fact that the molecular factors underlying protein function are often poorly characterized, task-specific, and highly context-sensitive. This makes it difficult to determine which features are likely to be informative for a given predictive objective. Furthermore, these hand-made representations lack contextual awareness: Each amino acid is encoded independently of its sequence or structural surroundings, making it difficult to capture cooperative effects or long-range dependencies. In contrast, recent deep learning approaches generate rich contextual embeddings that adapt to the position and environment of residues within full sequences or structures. These data-driven models reduce the need for manual feature design and have shown improved generalization across a wide range of protein-related tasks.
\section{Sequence-Based Approaches}
\label{sec:sequence}

Protein sequences, composed of linear chains of amino acids, represent the primary structure of proteins and encode the fundamental information that determines their higher-order structures and functions. These sequences can be viewed as a ``biological language,'' where the order and context of amino acids carry implicit rules and patterns. Inspired by this analogy, sequence-based protein representation learning (PRL) methods adapt techniques from natural language processing to learn the statistical ``grammar'' embedded within sequences. By training on large-scale sequence data, these models map proteins into latent vector spaces, where the resulting embeddings reflect meaningful biological relationships. Sequence-based approaches can be broadly categorized into aligned methods, which incorporate evolutionary information through Multiple Sequence Alignments (MSAs), and non-aligned methods, which learn directly from individual sequences.

\subsection{Aligned Sequence Approaches}

Aligned sequence methods leverage evolutionary relationships by analyzing Multiple Sequence Alignments (MSAs), which align amino acid sequences of homologous proteins from different organisms to identify conserved residues. An MSA arranges these sequences into a matrix where each row represents a protein and each column corresponds to a specific residue position, enabling systematic comparison of residue variability and conservation. The core insight behind MSAs is that conserved positions often reflect evolutionary constraints, such as structural stability, enzymatic activity, or interaction specificity, since mutations at these sites are typically deleterious and thus selected against. Consequently, MSAs encode rich contextual information that captures residue co-evolution, functional motifs, and structural dependencies, making them a powerful tool for studying protein behavior.

This evolutionary signal has made MSAs a valuable input for computational models aimed at inferring protein structure and function. Recent work has successfully used MSAs to learn biologically meaningful representations by modeling co-evolutionary patterns and conserved sequence features \cite{rao2021msa, jumper2021highly, zheng2024improving, chen2024msagpt}. Among these, two of the most influential and foundational are the MSA Transformer and AlphaFold. MSA Transformer \cite{rao2021msa} is a representation learning model that processes MSAs using axial attention, a mechanism that applies self-attention along both the rows (sequences) and columns (residue positions) of the alignment. This allows the model to jointly capture patterns of conservation and variation across homologous sequences, resulting in expressive protein embeddings grounded in evolutionary context. AlphaFold \cite{jumper2021highly}, although not a representation learning model in the strict sense, achieved a major breakthrough in structure prediction by deeply integrating MSA-derived information. Its Evoformer module refines MSA and pairwise residue representations through a series of attention-based communication blocks, enabling the model to infer spatial and functional constraints with remarkable accuracy. These two models are among the earliest and most widely recognized for their effective use of MSA data, and they have laid the foundation for many subsequent advances in protein representation learning and structure prediction.

Despite their effectiveness, MSA-based approaches have notable limitations. Their success depends on the availability of homologous sequences, which may be limited or absent for orphan proteins or de novo-designed sequences \cite{10.1093/bib/bbad217}. Moreover, constructing high-quality MSAs can be computationally expensive, especially at scale. These constraints have prompted increasing interest in non-aligned sequence methods that learn directly from individual sequences without relying on explicit evolutionary data \cite{lin2023evolutionary}.

\subsection{Non-Aligned Sequence Approaches}

Non-aligned sequence methods learn directly from individual protein sequences without relying on Multiple Sequence Alignments (MSAs) or explicit evolutionary context. These approaches are particularly valuable for proteins lacking sufficient homologs and are well suited for large-scale modeling, where the construction of MSAs can be computationally infeasible. In contrast to MSA-based methods, which explicitly leverage homologous sequences to capture evolutionary conservation, non-aligned models learn implicit biological constraints from the statistical co-occurrence of residues across large and diverse sequence corpora. This allows them to generalize more flexibly to proteins with limited or no evolutionary annotations, including those from underrepresented families or synthetic origins.

Deep learning has enabled a shift toward data-driven methods that learn representations from raw sequences. Early neural models such as Variational Autoencoders (VAEs) \cite{sinai2017variational, ding2019deciphering} captured the generative structure of protein space by mapping sequences to low-dimensional latent vectors. Recurrent architectures such as RNN and LSTM further advanced this line of work by modeling long-range sequence dependencies \cite{almagro2020language}. A notable milestone was UniRep \cite{alley2019unified}, which introduced a self-supervised approach using mLSTM trained on millions of sequences. UniRep demonstrated that fixed-length embeddings could predict a range of biochemical properties, laying the foundation for the era of protein language models (PLMs).

More recently, transformer-based Protein Language Models (PLMs) have become the dominant paradigm, drawing inspiration from large language models in natural language processing. Models such as the Evolutionary Scale Modeling (ESM) series \cite{rives2021biological, lin2023evolutionary, hayes2024simulating}, ProtTrans \cite{elnaggar2021prottrans}, ProteinBERT \cite{brandes2022proteinbert}, and ProLLaMA \cite{lv2024prollama} are pre-trained on massive protein sequence datasets using masked language modeling objectives. During training, PLMs learn contextual dependencies by predicting masked amino acids, resulting in internal embeddings that encode rich, task-agnostic information about sequence structure and function. These representations form a learned latent space, where similar proteins are embedded closer together, enabling knowledge transfer across diverse sequence types. In addition to general-purpose PLMs, specialized models have been developed to target particular protein classes. For example, antibody-specific PLMs \cite{olsen2022ablang, gao2023pre, shuai2023iglm, kenlay2024large} and enzyme-focused models \cite{munsamy2022zymctrl} incorporate domain-specific inductive biases to improve performance in tasks such as affinity maturation and prediction of catalytic function.
\section{Structure-Based Approaches}
\label{sec:structure}

\begin{figure}[ht]
\centering
\includegraphics[width=\columnwidth]{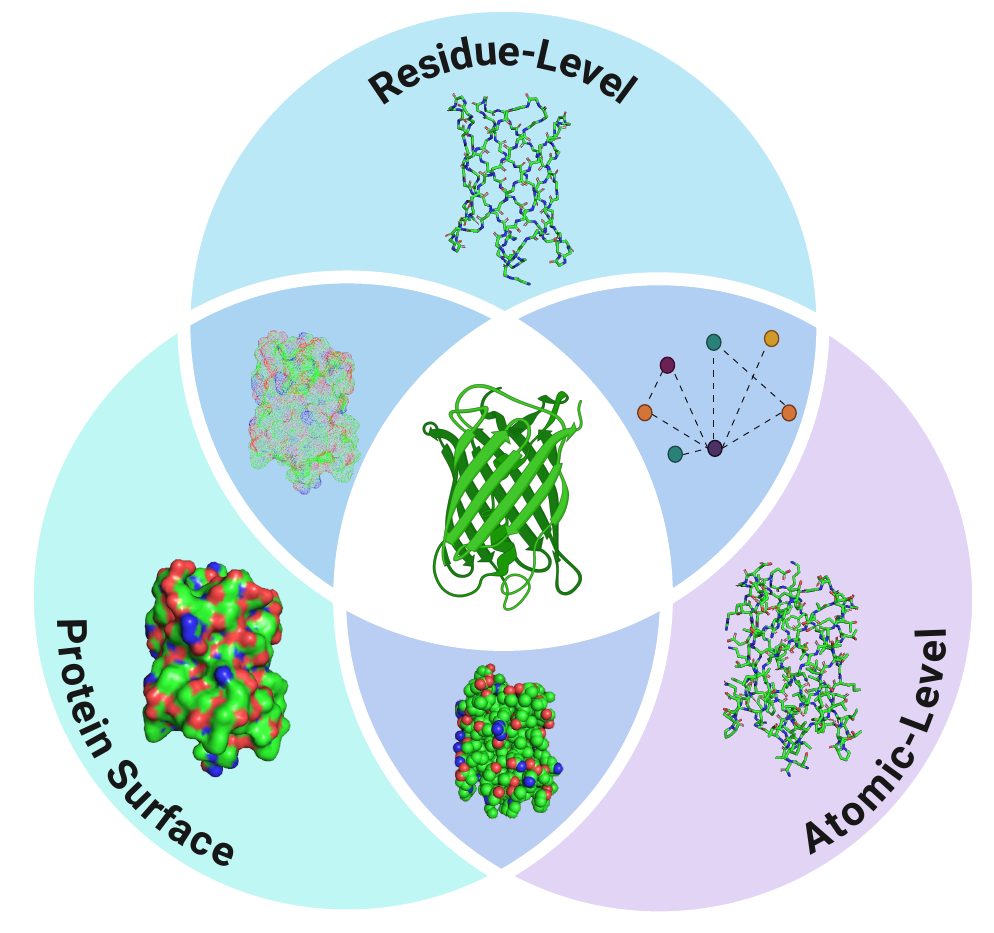}
\vspace{-0.25cm}
\caption{\label{fig:structure_based} Illustration of different structural representations of a protein at the tertiary level, along with their typical computational representations. Residue-Level Representation models the protein backbone, typically using alpha carbon (C$\alpha$) atoms to capture the overall fold and residue connectivity, with a corresponding graph-based representation. Atomic-Level Representation considers all individual atoms within the protein structure, including backbone and side-chain atoms, commonly represented as a point cloud or an all-atom graph that captures atomic interactions. Protein Surface Representation focuses on the solvent-accessible surface, highlighting geometric and physicochemical properties that influence binding and molecular recognition, often modeled using a surface mesh or a point cloud representation that encodes local curvature and electrostatic potential.}
\end{figure}

Structure-based approaches capture the three-dimensional organization of proteins, which is crucial to understanding their structural and functional properties. These approaches leverage structural data at various levels of granularity, as illustrated in \cref{fig:structure_based}, including atoms, residues, and surfaces, to provide richer insight into protein stability, function, and interactions. In addition, they integrate symmetry principles to improve learning efficiency and improve representation quality.

\subsection{Residue-Level Representation}
Residue-level representation models focus on capturing structural information at the amino acid level. Although sequence-based representations are limited to linear order, residue-level approaches consider the spatial arrangement of residues in the folded protein. This is particularly important because residues that are distant in the primary sequence may be brought into close proximity in the three-dimensional structure due to protein folding. By explicitly modeling these spatial relationships, residue-level representations enable a more comprehensive understanding of protein stability, interactions, and functional properties.

Several notable studies have advanced residue-level representations by leveraging graph-based approaches \cite{hermosilla2022contrastive, xia2022fast, zhou2022lightweight, le2022representation, guo2022self,  wang2023learning, chen20233d, zhang2023protein, Gao2023, voitsitskyi20233dprotdta}. In this representation, proteins are modeled as graphs, where the nodes correspond to residues, typically represented by the alpha carbon ($C\alpha$) of each amino acid along with its 3D coordinates. Node features frequently include amino acid type, physicochemical properties (e.g. charge, hydrophobicity, and polarity) and secondary structure annotations, providing rich information about each residue's role and environment. The edges of the residue-level graphs vary to reflect different relationships between residues \cite{zhang2023protein}:
\begin{itemize}
    \item \textbf{Sequential edges} connect residues that are adjacent in the primary sequence, capturing the linear backbone connectivity of the protein chain.
    \item \textbf{Radius edges} link residues within a predefined distance cut-off in 3D space, emphasizing local spatial neighborhoods critical for structural integrity.
    \item \textbf{K-nearest neighbor edges} connect each residue to its closest spatial neighbors, independent of their sequence order, effectively capturing geometric relationships across the protein.
\end{itemize}
Additionally, edge features may encode interaction types, such as hydrogen bonds, disulfide bonds, or hydrophobic interactions, as well as geometric properties like bond angles and dihedral angles, further enhancing the model’s ability to capture structural and functional relationships.

Beyond graph-based approaches, recent work has explored discretized residue-level representations for scalable structural modeling and comparison \cite{van2024fast, zhang2024balancing, gaujac2024learning}. These methods encode local or global 3D structural environments into a finite set of tokens, effectively transforming protein structures into symbolic sequences. Many of these representations are learned using vector-quantized variational autoencoders (VQ-VAEs), which cluster local residue geometries into discrete structural alphabets. A typical and widely adopted example is Foldseek \cite{van2024fast}, which constructs such an alphabet by assigning each residue a token that summarizes its spatial context relative to neighboring residues. This enables sequence-style processing with efficient alignment while retaining biologically relevant structural information. Unlike models that explicitly define spatial graphs, discretized representations offer compact and alignment-friendly abstractions suitable for large-scale structure analysis and modeling.

\subsection{Atomic-Level Representation}

Atomic-level representation models capture the fine-grained structural and chemical characteristics of proteins by explicitly modeling individual atoms and their spatial arrangements. This level of granularity preserves critical details necessary to understand molecular interactions, since it includes all atoms within a protein, both backbone and side chains, rather than approximating each residue with a single representative point, such as alpha carbon ($C\alpha$). By incorporating atom-specific features and spatial relationships, atomic-level approaches provide a richer and more precise encoding of the protein structure. Several modeling paradigms have been proposed at this level, including graph-based representations, point cloud methods, and voxel-based 3D convolutional neural networks (3D CNNs), each offering unique inductive biases and trade-offs in terms of expressiveness, efficiency, and spatial resolution.

Graph-based methods represent protein structures as atom-level graphs, where nodes correspond to individual atoms annotated with features such as element type, partial charge, hybridization state, or residue identity \cite{wang2023learning, wu2023atomic, jiale2024pretraining}. Edges define pairwise relationships between atoms, typically based on chemical bonds, spatial proximity, or non-covalent interactions. A common strategy is to use $k$-nearest neighbor (KNN) connections to flexibly capture both bonded and non-bonded atomic interactions. These graphs are well-suited for message-passing neural networks and geometric learning architectures that propagate information over the molecular graph to model structural dependencies.

Point cloud-based methods treat atoms as an unordered set of points in three-dimensional space, each point associated with features such as atom type or partial charge \cite{wang2022point, nguyen2024multimodal}. These approaches forego explicit graph connectivity, relying instead on spatial convolutions or attention mechanisms to learn directly from the geometric distribution of atoms. This makes point cloud representations efficient and adaptable, especially for modeling flexible or irregular structures, where fixed connectivity may be limiting. The raw spatial encoding also allows these models to generalize across diverse molecular conformations.

Voxel-based 3D convolutional neural networks (3D CNNs) provide a dense grid-based representation of atomic environments \cite{torng2019high, torng20173d}. In this framework, atoms are projected into a fixed-resolution three-dimensional lattice, where each voxel encodes the presence and chemical identity of atoms within a defined spatial window. These volumetric grids are processed using 3D convolutional layers that capture local geometric and chemical patterns. Unlike graph- and point-cloud models, 3D CNNs operate in structured input spaces, leveraging established computer vision techniques. While computationally more demanding due to their dense spatial encoding, 3D CNNs offer high spatial fidelity and do not require predefined topological constraints.

\subsection{Protein Surface Representation}
A protein’s molecular surface is a compact, smooth boundary formed by the outermost atoms, exhibiting distinct chemical and geometric patterns. As the primary interface for molecular interactions, the protein surface plays a crucial role in molecular recognition, ligand binding, and protein-protein interactions. The protein surface representation focuses on modeling this exterior, where these interactions predominantly occur, providing a functionally relevant perspective that complements residue-level and atomic-level representations.

Several notable studies have advanced protein surface representations by using 3D mesh-based approaches \cite{gainza2020deciphering, riahi2023surface, mallet2023atomsurf, randolph2024invariant, marchand2025targeting} and point cloud-based approaches \cite{Sverrisson_2021_CVPR, liu2021octsurf, Srinivasan2022, sun2023dsr}. In mesh-based models, the protein surface is represented as a triangulated mesh, where the nodes correspond to discrete surface points, and the edges define geometric relationships between neighboring points, preserving the overall topology of the surface. These models effectively capture geometric features such as curvature, solvent accessibility, and surface normals, as well as chemical properties such as electrostatic potential and hydrophobicity. Due to their structured nature, mesh-based representations excel in tasks that require fine-grained surface characterization, such as functional site identification, enzyme active site modeling, and molecular docking orientation prediction. However, mesh-based approaches often rely on precomputed geometric and chemical features, making them computationally demanding and less adaptable for large-scale datasets.

In contrast, point-cloud-based approaches offer a more computationally efficient alternative by avoiding the need for precomputed features. These models operate directly on the raw set of atoms composing the protein, generating a point-cloud representation of the surface on the fly. Unlike mesh-based methods, point cloud approaches learn task-specific geometric and chemical features dynamically and apply convolutional operators that approximate geodesic coordinates in the tangent space. This eliminates the need for an explicit surface mesh and significantly reduces memory usage and computational overhead, making point cloud-based representations well-suited for large-scale protein-ligand binding predictions, protein-protein interface modeling, and high-throughput interaction analysis.

\subsection{Symmetry-Preservation and Equivariance in Structural Representation}

Incorporating three-dimensional (3D) structural information into PRL presents the challenge of ensuring that models correctly handle spatial transformations such as rotations and translations. Models relying on raw 3D coordinates risk developing spurious dependencies on absolute positioning, leading to inconsistent predictions for identical structures in different orientations. This limitation is particularly problematic in tasks such as protein folding, molecular docking, and prediction of binding sites, where the relative spatial geometry governs the molecular behavior. Symmetry-preserving and equivariant models address this issue by explicitly incorporating the inherent rotational and translational symmetries of protein structures, improving model performance, generalization, and computational efficiency. Such models ensure either invariance (i.e. output representation remains unchanged regardless of input transformations) or equivariance (i.e. output representation transforms accordingly to input transformations), preserving relative spatial relationships.

Recent advances in geometric deep learning have led to the development of rotation-equivariant \cite{pmlr-v139-satorras21a, NEURIPS2020_15231a7c, NEURIPS2019_03573b32} and permutation-equivariant graph neural networks \cite{maron2018invariant, 10.1063/1.5024797}, which explicitly encode spatial symmetries, improving PRL \cite{zhou2022lightweight, le2022representation, wu2023atomic, chen20233d}. By inherently preserving geometric properties, these models enhance the accuracy of predictions in tasks such as protein structure modeling and molecular interactions. Moreover, by eliminating the need for extensive data augmentation, such as generating rotated versions of protein structures, they reduce computational overhead while improving model robustness and efficiency.

In parallel, invariant models address spatial symmetries by learning transformation-independent features, such as interatomic distances, bond angles, and dihedral angles of the backbone \cite{Srinivasan2022, guo2022self, randolph2024invariant, cheng2024zero}. These geometric descriptors remain constant under rigid-body transformations, ensuring that identical protein structures in different orientations yield the same representation. For instance, atomic distances, bond angles, and dihedral angles capture essential spatial relationships while being inherently independent of absolute positioning. This approach simplifies learning, reduces computational complexity, and improves generalization across diverse protein structures. By eliminating orientation and positional variance, invariant models streamline training while maintaining high predictive performance.
\section{Multimodal-Based Approaches}
\label{sec:multimodal}

Multimodal-based approaches aim to integrate multiple sources of protein data, such as sequences, structures, and functional annotations, to create richer and more comprehensive representations. Each modality captures distinct aspects of a protein's characteristics: sequences provide information on the amino acid composition and evolutionary history, structures reveal the three-dimensional conformation critical for function, and functional annotations describe biological roles, molecular interactions, and phenotypic effects. By combining these complementary data types, multimodal approaches address the limitations inherent in single-modality models, leading to more accurate, robust, and generalizable protein representations.

\subsection{Integration of Sequence and Structure Representations}

One of the most prominent directions in multimodal PRL involves the integration of sequence and structure data. While Protein Language Models (PLMs) have demonstrated remarkable success in capturing evolutionary and contextual information from large-scale sequence datasets, their capability to encode the three-dimensional structural context essential for protein function remains limited. This limitation arises because sequence-only PLMs primarily model linear dependencies and lack explicit information about the spatial arrangement of residues. To address this gap, researchers have developed three primary strategies for incorporating structural information into PLMs to enhance their ability to model protein functions and interactions:

\begin{enumerate}
\item The first approach focuses on the integration of sequence-derived information into structure-based models. In this method, PLMs are used to extract amino acid embeddings from sequences, which are then treated as node features on residue-level graphs \cite{wang2022lm, gligorijevic2021structure, wu2023integration, blaabjerg2024ssemb, ahmed2025improved}. These graphs are constructed using structural data, capturing spatial dependencies between residues through edges defined by proximity, interaction types, or geometric constraints. This strategy takes advantage of the contextual richness of PLMs while integrating the representations in the three-dimensional structure of the protein, resulting in more comprehensive and informative models.

\item The second approach takes the reverse direction by integrating structural information directly into the training of PLMs. Instead of relying solely on amino acid sequences, this method introduces structural data, such as residue distances, angles, or spatial neighbors, into the input or learning objective of the model \cite{zheng2023structure, yang2023masked, su2024saprot, sun2024structure, li2025large}. In some cases, models replace or augment traditional sequence tokens with residue-level structural tokens that describe the 3D context of each amino acid. These tokens, often derived from geometric relationships or learned structural alphabets, allow the model to process structural patterns using the same language-modeling techniques applied to sequences. By learning both the sequence and the structure at the token level, these models can generate more informative and biologically grounded representations.

\item The third approach treats sequence and structural data as distinct modalities, encoding them separately using models tailored to each \cite{lee2023pre, wang2023s, cheng2024zero, zhang2024pre, barton2024enhancing, ngo2024multimodal, nguyen2024multimodal, liu2025rna}. The resulting representations are then fused through techniques such as attention mechanisms, contrastive learning, or embedding concatenation. This multi-branch strategy allows models to learn complementary features from both modalities while preserving the unique contributions of each, leading to enhanced representation quality and generalizability.
\end{enumerate}

\subsection{Integration of Sequence and Functional Representations}

While Protein Language Models (PLMs) have advanced the understanding of sequence patterns, they often lack the explicit biological knowledge required to accurately capture protein function. Sequence data alone provide limited information on a protein's specific roles or its molecular interactions within cellular environments, which can hinder model performance in tasks such as prediction of protein functions, protein contact mapping, and protein-protein interaction analysis. Given that a protein’s structure fundamentally determines its function, models can be significantly improved by incorporating functional knowledge that highlights similarities among proteins with comparable shapes or sequences. Functional annotations provide comprehensive insights into the biological processes, molecular functions, and cellular components of a protein. Integrating such functional data into PLMs enriches protein representations, allowing models to more effectively generalize across diverse biological tasks and applications.

To address the limitations of sequence-only models, functional data are typically treated as a separate modality. Researchers employ biomedical language models such as BioBERT \cite{lee2020biobert} and PubMedBERT \cite{gu2021domain}, which are trained in large-scale biomedical literature and ontologies, to generate functional embeddings that capture rich contextual and domain-specific knowledge. These functional embeddings are then fused with sequence-derived representations using methods such as contrastive learning or joint embedding frameworks \cite{zhang2022ontoprotein, zhou2023protein, xu2023protst, kilgore2025protein}. This multimodal integration enhances the representational capacity of PLMs, enabling more accurate and biologically informed predictions on a range of functional and interaction-based tasks.

\subsection{Unified Representations of Sequence, Structure, and Function}

Unified representations are used to simultaneously integrate sequence, structural, and functional data into a cohesive modeling framework, providing a comprehensive view of protein characteristics. Although sequence-based models capture evolutionary and contextual relationships and structure-based models provide insights into three-dimensional conformation, the integration of functional annotations adds a critical biological context regarding the roles and interactions of a protein within cellular systems. By combining these modalities, unified representations offer a holistic understanding of proteins, allowing models to capture the complex interaction between a protein's linear sequence, its folded structure, and its biological function.

To achieve this, researchers typically adopt multi-branch architectures, where each modality (sequence, structure, and function) is encoded separately using specialized models. The output of these models is then fused using techniques such as attention mechanisms, contrastive learning, or cross-modal embedding fusion \cite{hu2023multimodal, abdine2024prot2text}. This approach ensures that the unique contributions of each modality are preserved while enabling the model to learn complementary features that improve the overall quality of the representation. By capturing the full complexity of protein biology, these multimodal approaches pave the way for more accurate, generalizable, and biologically meaningful models.
\section{Complex-Based Approaches}
\label{sec:complex}

Complex-based approaches focus on learning representations of entire protein complexes, capturing the structural and functional relationships between interacting molecules. These methods model interactions between proteins and ligands or protein-protein assemblies, which are fundamental to biological processes such as enzyme activity, signal transduction, and molecular recognition. By explicitly representing protein complexes, these approaches aim to learn context-aware embeddings that encode both individual molecular properties and interaction-specific features, facilitating tasks such as binding affinity prediction, molecular docking, and complex structure generation.

\subsection{Protein-Ligand Complex Representation}

The protein-ligand complex representation aims to learn a unified representation that effectively captures the interactions between proteins and small molecules, enabling the models to better understand binding mechanisms and affinity relationships. Rather than representing the entire protein structure, these approaches typically focus on the binding pocket, the localized region where molecular interactions occur. Upon concentration on this functionally relevant site, models can capture more accurately the spatial, chemical, and sequence-dependent features that govern ligand binding and molecular recognition. Approaches for protein-ligand complex representation can be broadly classified into two following categories, differentiated by their use of 3D structural information. 

The first category follows a dual-encoder framework \cite{karimi2019deepaffinity, wang2022electra, pei2022smt, gao2023co, nakata2023end, nguyen2024proteinrediff, gao2024drugclip}, where proteins and ligands are encoded separately, without explicitly modeling their spatial interactions. Protein representations are commonly derived from protein language models (PLMs) or graph-based encoders, while ligand representations are generated using SMILES-based transformers, molecular fingerprints, or graph neural networks (GNNs). Since these models do not require pre-existing 3D structural data for the full complex, they are particularly advantageous for high-throughput virtual screening and cases where structural information is unavailable. During fine-tuning, a lightweight interaction module is introduced to align and refine these representations for downstream tasks such as binding affinity prediction and molecular docking. This modular approach improves scalability and facilitates transfer learning across diverse protein-ligand datasets.  

The second category focuses on fine-grained interaction modeling, where 3D structural information of the complete protein-ligand complex is explicitly incorporated to learn atomic- or residue-level interactions. Unlike dual-encoder approaches, which align representations post hoc, these methods directly encode spatial and chemical dependencies within the binding pocket, capturing fine-grained molecular interactions. The binding mechanism of protein-ligand complexes is highly intricate, involving a diverse array of non-covalent forces such as $\pi$-stacking, $\pi$-cation interactions, salt bridges, water bridges, hydrogen bonds, hydrophobic interactions, and halogen bonds \cite{C7MD00381A}. Previous studies \cite{zheng2019onionnet, moon2022pignet, feng2024proteinligand} have demonstrated that explicitly modeling these atomic-level forces significantly improves predictive performance, underscoring the importance of 3D molecular force modeling in PRL.

\subsection{Protein Complex Representation}

Protein complexes consisting of two or more interacting polypeptide chains are fundamental to biological processes such as signal transduction, immune response, enzymatic activity, and molecular assembly. These complexes exhibit significant diversity in size, stability, and interaction mechanisms, ranging from small transient interactions to large, stable macromolecular structures such as ribosomes, chaperones, and antibody-antigen complexes. The representation of protein complexes closely follows the methodology used in protein-ligand complex modeling, where the goal is to learn a joint representation that captures both individual protein properties and interaction-specific features. Approaches to this problem can be broadly categorized on the basis of whether they explicitly encode the full complex structure or rely on separate representations of individual proteins.  

Structure-free approaches, which encode individual proteins separately before aligning them in a shared representation space, are widely applied to various challenges in protein complex modeling \cite{yuan2023dg, li2024mvsf, bandara2024deep}. These approaches do not explicitly model the structural constraints of the full complex during encoding. Instead, interactions are inferred post hoc through mechanisms such as contrastive learning or learned scoring functions. Among the many tasks tackled by these approaches, the prediction of the structure of the protein complex, the determination of the spatial arrangement of the interacting polypeptides given only sequence or monomeric structural information, is one of the most critical \cite{evans2021protein, giulini2023towards, ruffolo2023fast}. 

Structure-aware approaches, on the other hand, explicitly encode the entire 3D structure of the protein complex, incorporating spatial and physicochemical constraints to model binding interfaces and interaction dependencies \cite{reau2023deeprank, yue2024integration, wang2024eggnet}. These methods leverage geometric deep learning, graph-based modeling, and energy-based optimization techniques to refine complex formation by capturing atomic-level interactions and physical constraints. Because these approaches explicitly encode the joint structural context of interacting proteins, they can provide more interpretable predictions, improve docking accuracy, and enhance the design of protein-protein interactions for therapeutic applications. However, they require high-quality structural data, which may not always be available, and must balance computational efficiency with accuracy when handling large macromolecular assemblies.

\begin{table*}
\centering
\caption{Comparative Summary of Strengths and Limitations of PRL Approaches.}
\label{tab:prl_summary}
\resizebox{\textwidth}{!}{
\begin{tabular}{p{0.15\textwidth}p{0.42\textwidth}p{0.42\textwidth}}
\toprule
\textbf{PRL Approach} & \textbf{Strengths} & \textbf{Limitations} \\
\midrule
\textbf{Feature-Based} &
\begin{itemize}[noitemsep,topsep=0pt]
    \item Simple and interpretable.
    \item Leverages well-established biochemical properties.
    \item Compatible with traditional machine learning models.
\end{itemize} &
\begin{itemize}
    \item Requires manual feature selection.
    \item May oversimplify protein characteristics.
    \item Struggles to generalize across diverse protein-related tasks.
\end{itemize} \\  
\midrule
\textbf{Sequence-Based} & 
\begin{itemize}
    \item Utilizes large-scale protein sequence data for training Protein Language Models (PLMs).
    \item Captures evolutionary and contextual sequence information.
\end{itemize} &
\begin{itemize}
    \item MSA-based approaches require homologous sequences and can be computationally expensive.
    \item Non-aligned methods may lack evolutionary context.
\end{itemize} \\  
\midrule
\textbf{Structure-Based} & 
\begin{itemize}
    \item Encodes rich spatial and functional information.
    \item Symmetry-aware models improve generalization.
\end{itemize} &   
\begin{itemize}
    \item Requires high-quality structural data.
    \item Computationally intensive, especially at the atomic level.
    \item Surface-based representations may be memory-intensive.
\end{itemize} \\  
\midrule
\textbf{Multimodal-Based} &  
\begin{itemize}
    \item Integrates multiple data sources for comprehensive protein representations.
    \item Improves model accuracy and generalization.
\end{itemize} &  
\begin{itemize}
    \item Requires large, high-quality multimodal datasets.
    \item Increases computational complexity.
    \item Challenges in effectively fusing heterogeneous data modalities.
\end{itemize} \\  
\midrule
\textbf{Complex-Based} &  
\begin{itemize}
    \item Captures interaction-specific features in protein-ligand and protein-protein complexes.
\end{itemize} &
\begin{itemize}
    \item Requires high-quality structural data.
    \item Structure-free methods may lack spatial constraints.
\end{itemize} \\  
\bottomrule
\end{tabular}}
\end{table*}
\section{Databases for Protein Representation Learning}
\label{sec:databases}

Large-scale biological databases are essential for Protein Representation Learning (PRL), which offers the sequence, structure, and functional data required for model development and evaluation. These resources have not only enabled large-scale pretraining, but have also shaped the design and evaluation of diverse representation methods. Table~\ref{tab:protein_databases} summarizes key databases in different data modalities, reflecting their scale and utility in modern PRL research. Despite the breadth of available data, there are significant imbalances between modalities. Sequence data are highly abundant, with repositories such as UniProtKB and MGnify containing hundreds of millions to billions of entries, providing a rich foundation for capturing evolutionary and biochemical patterns in proteins. In contrast, experimentally resolved protein structures remain relatively limited, as databases such as the Protein Data Bank (PDB) are constrained by the time, cost, and complexity of structural determination techniques. The emergence of structure prediction resources, including AlphaFold DB and ESMAtlas, has significantly increased structural coverage, supporting the development of structure-aware models. However, these predicted models may vary in accuracy and often lack experimental validation, particularly in regions of intrinsic disorder or low confidence. Functional annotations represent an even sparser modality: Only a small subset of known proteins are associated with experimentally confirmed functions, and many annotations are computationally inferred, introducing varying degrees of uncertainty. These imbalances pose a particular challenge for multimodal approaches that rely on aligned and balanced data across sequence, structure, and function.

\begin{table*}[ht]
    \centering
    \caption{Summary of widely used databases in Protein Representation Learning (PRL), categorized by sequence, structure, and function.}
    \resizebox{\textwidth}{!}{
    \begin{tabular}{
        >{\raggedright\arraybackslash}p{0.18\textwidth}
        >{\raggedright\arraybackslash}p{0.10\textwidth}
        >{\raggedright\arraybackslash}p{0.12\textwidth}
        >{\raggedright\arraybackslash}p{0.60\textwidth}
    }
        \toprule
        \textbf{Name} & \textbf{Data Type} & \textbf{Scale} & \textbf{Description} \\
        \midrule
        UniProtKB \cite{boutet2007uniprotkb, EDITtoTrEMBL} & Sequence, Function & $\sim$252M & Comprises Swiss-Prot (manually curated) and TrEMBL (automatically annotated); widely used in pretraining and function annotation. \\
        UniRef \cite{suzek2007uniref} & Sequence & $\sim$250M & Clusters UniProt sequences at different identity thresholds to reduce redundancy and improve training efficiency. \\
        Pfam \cite{mistry2021pfam} & Sequence & $\sim$22K & Defines protein families and domains using MSAs and profile Hidden Markov Models to support motif-level annotation. \\
        MGnify \cite{richardson2023mgnify} & Sequence & $\sim$2.4B & Provides non-redundant protein sequences predicted from metagenomic environmental samples to increase diversity. \\
        PDB \cite{berman2000protein} & Structure & $\sim$231K & Repository of experimentally determined protein structures via X-ray crystallography, NMR, and cryo-EM. \\
        AlphaFold DB \cite{varadi2024alphafold} & Structure & $\sim$214M & Stores high-confidence predicted protein structures from AlphaFold to improve structural coverage. \\
        ESMAtlas \cite{lin2023evolutionary} & Structure & $\sim$772M & Contains predicted structures from ESMFold, including many novel and uncharacterized proteins. \\
        Gene Ontology \cite{gene2023gene} & Function & $\sim$45M& Provides structured functional annotations across biological processes, molecular functions, and cellular components. \\
        \bottomrule
    \end{tabular}
    }
    \label{tab:protein_databases}
\end{table*}

\section{Applications}
\label{sec:applications}
\begin{figure*}[ht]
\centering
\includegraphics[width=\textwidth]{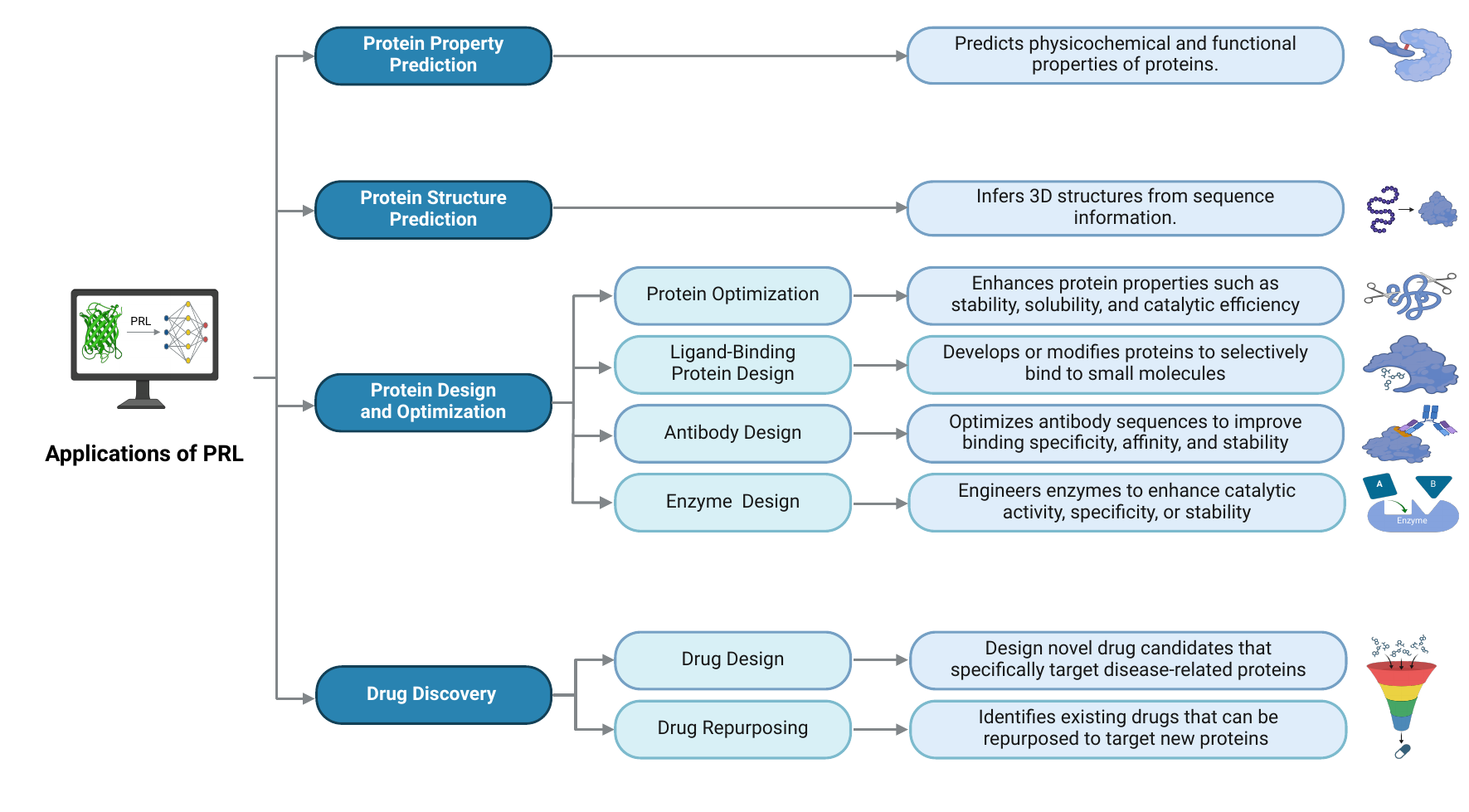}
\vspace{-0.25cm}
\caption{\label{fig:applications} Overview of key applications of Protein Representation Learning (PRL). PRL enables advancements in multiple domains, including protein property prediction, structure prediction, protein design and optimization, and drug discovery. }
\end{figure*}

Protein Representation Learning (PRL) has revolutionized biological research and biomedical innovation, providing data-driven solutions to understand and engineering proteins. By learning meaningful representations of protein sequences, structures, and interactions, PRL can enable a wide range of applications, from predicting fundamental protein properties to advancing drug discovery. As illustrated in \cref{fig:applications}, PRL applications can be broadly categorized into four main domains: prediction of protein properties, prediction of protein structure, design and optimization of proteins and drug discovery.

In the following sections, each application area is presented through a structured discussion that includes: (i) a problem statement defining the task, (ii) an explanation of its significance in biological and biomedical contexts, (iii) an overview of commonly used benchmarks for evaluating model performance, and (iv) a discussion of models that have advanced the field, with observations on their representation learning strategies. This structure provides a comprehensive and comparative perspective on how PRL is shaping advancements in protein science and biopharmaceutical applications.

\subsection{Protein Property Prediction}

Predicting protein properties involves estimating key characteristics such as solubility, stability, enzymatic activity, and binding affinity from sequence or structural data. These properties are fundamental to protein function and have significant implications in biotechnology, medicine, and drug discovery. Accurate prediction models reduce experimental costs and accelerate the design of novel therapeutics and industrial enzymes, making protein property prediction a central task in PRL.

Benchmarks for evaluating protein property prediction models are categorized according to the type of property that is being assessed. For individual protein properties, Tasks Assessing Protein Embeddings (TAPE) \cite{rao2019evaluating} provides a standardized evaluation framework, including fluorescence prediction and protease stability assessment. Another widely used benchmark, UniProt \cite{boutet2007uniprotkb}, serves as a comprehensive database of protein sequences, functional annotations, and metadata from various organisms, offering a valuable resource for assessing sequence-based property predictions in various protein families and functional contexts. In contrast, interaction-based properties, such as protein-ligand and protein-protein interactions, require different benchmarks that incorporate structural and binding information. Prominent datasets in this category include PDBBind \cite{PDBbind}, Leak Proof PDBBind \cite{li2024leakproofpdbbindreorganized}, Davis \cite{Davis}, KIBA \cite{KIBA}, PLINDER \cite{Durairaj2024.07.17.603955} and PINDER \cite{kovtun2024pinder}, which evaluate models based on binding affinity predictions. These datasets serve as widely recognized standards, enabling consistent model comparison and performance assessment across different protein property prediction tasks.

With the remarkable advancements in Protein Language Models (PLMs), most current research leverages these models to extract meaningful representations of protein sequences, applying them to a wide range of property prediction tasks. These models have been successfully employed for both individual protein properties \cite{yang2018learned, yu2023enzyme, buton2023predicting} as well as interaction-based properties \cite{yuan2025mutualdata,  yuan2023dg, ha2025lantern}. Beyond sequence-based representations, many efforts have explored the integration of multiple modalities, such as structural, physicochemical, and functional data, to further enhance protein representations beyond what PLM alone can achieve \cite{wang2022lm, nguyen2024multimodal, ngo2024multimodal, luo2024enhancing}. By incorporating additional biological information, these approaches aim to provide a more comprehensive view of protein function and interactions, particularly in tasks where spatial and contextual features play a significant role, such as protein-ligand binding and protein-protein interactions.

\subsection{Protein Structure Prediction}

Protein structure prediction, also known as protein folding, aims to determine the three-dimensional conformation of a protein from its amino acid sequence. Since protein structure dictates function, accurately predicting spatial arrangements is essential for understanding molecular interactions, drug binding, and enzymatic activity. Traditionally, protein structures have been determined using experimental techniques such as X-ray crystallography \cite{https://doi.org/10.1111/febs.12796} and cryo-electron microscopy (Cryo-EM) \cite{BAI201549}, which provide high-resolution structural insights. However, these methods are often time-consuming, costly, and limited by challenges such as crystallization feasibility and sample preparation \cite{venien2017cryo}. Advances in computational structure prediction have transformed the field by providing faster, scalable alternatives that complement experimental approaches, broadening the scope of protein engineering and biomedical research.

To evaluate the accuracy of structure prediction models, several benchmarking platforms have been established. The Critical Assessment of Techniques for Protein Structure Prediction (CASP) \cite{kryshtafovych2021critical}, a biennial blind assessment, is widely regarded as the gold standard for evaluating protein folding models. In addition, Continuous Automated Model EvaluatiOn (CAMEO) \cite{robin2021continuous} provides real-time assessments by continuously comparing predictions against newly released experimental structures. These benchmarks are instrumental in validating and refining computational models, ensuring their reliability and applicability in structural biology.

Protein structure prediction models can be broadly categorized into MSA-based and MSA-free approaches. MSA-based methods leverage multiple sequence alignments to capture evolutionary relationships, often resulting in higher accuracy for proteins with rich evolutionary data. Notable models in this category include AlphaFold \cite{jumper2021highly}, RosettaFold \cite{baek2021accurate}, OpenFold \cite{ahdritz2024openfold}, and MSAGPT \cite{chen2024msagpt}. In contrast, MSA-free methods predict structures directly from single sequences without relying on evolutionary context, making them more efficient and applicable to proteins with limited sequence homology. Examples of MSA-free models include ESMFold \cite{lin2023evolutionary}, OmegaFold \cite{wu2022high}, and HelixFold-Single \cite{fang2023method}. While these models have greatly improved protein structure prediction, they are primarily limited to static conformations, as they are trained on crystallographic and experimentally resolved structures. However, proteins are inherently dynamic and undergo conformational changes essential to their function. To address this limitation, recent advancements have introduced generative models capable of predicting conformational ensembles, such as BioEmu \cite{lewis2024scalable}.

\subsection{Protein Design and Optimization}

Protein design and optimization involve strategies for engineering proteins with specific structural and functional properties. Broadly, these strategies fall into two main categories: sequence-based design, which optimizes amino acid sequences to achieve target properties, and structure-based design, which leverages protein structural features either by generating backbones and applying inverse folding to identify compatible sequences or by co-designing sequence and structure to enhance functional and structural outcomes. These approaches have been applied to various protein types, including individual proteins with optimized functionality, ligand-binding proteins for molecular interactions, antibodies for therapeutic applications, and enzymes for improved catalytic efficiency in industrial and biomedical settings.

\subsubsection{Individual Protein Property Optimization}

Individual protein property optimization focuses on enhancing the intrinsic characteristics of a protein, such as stability, activity, specificity, and catalytic efficiency, without explicitly considering interactions with other molecules. Traditional approaches, such as directed evolution with random mutagenesis \cite{doi:10.1021/ar960017f, TRACEWELL20093, doi:10.1073/pnas.1215251110}, iteratively introduce mutations and select improved variants. While these methods have been widely used and have led to many successful optimizations, they often require extensive experimental screening. Computational strategies have been developed to facilitate a more systematic exploration of sequence space, offering alternatives that reduce experimental reliance while aiming to improve efficiency in protein engineering.

To evaluate optimization strategies, several benchmark datasets have been established for sequence-based optimization, each representing distinct challenges between protein targets, originating organisms, sequence lengths, dataset sizes, and fitness objectives. Notable benchmarks include Green Fluorescent Protein (avGFP) \cite{sarkisyan2016local}, Adeno-Associated Viruses (AAV) \cite{bryant2021deep}, TEM-1 $\beta$-Lactamase (TEM) \cite{firnberg2014comprehensive}, Ubiquitination Factor Ube4b (E4B) \cite{starita2013activity}, Aliphatic Amide Hydrolase (AMIE) \cite{Wrenbeck2017}, Levoglucosan Kinase (LGK) \cite{Klesmith2015}, Poly(A)-binding Protein (Pab1) \cite{Melamed_2013}, and SUMO E2 Conjugase (UBE2I) \cite{weile2017framework}. These datasets enable a standardized assessment of different protein optimization methods.

Earlier approaches focused on optimizing proteins in discrete sequence space. These methods explored mutation-based search strategies, reinforcement learning, and PLMs to propose beneficial sequence modifications \cite{emami2023plug, tran2024protein}. Building upon this foundation, recent advancements in PRL have introduced computational frameworks capable of mapping protein sequences into fitness landscapes. These models aim to capture sequence-function relationships and facilitate direct sequence optimization based on learned representations. Unlike experimental screening, PRL-based approaches rely on large-scale protein datasets to infer meaningful sequence features and predict fitness variations. A common strategy in fitness landscape modeling integrates sequence and fitness information within a machine learning framework. One such approach involves jointly training an autoencoder with a prediction network, where an encoder extracts sequence features, a decoder reconstructs sequences, and a separate model predicts fitness values from the latent space \cite{castro2022transformer, 10.1088/2632-2153/adc2e2}. Protein Language Models (PLMs) are often used as encoders to capture contextual sequence representations, and the prediction network guides latent space organization based on fitness variations.

Other studies have explored methods to improve fitness landscape modeling through smoothing techniques, aiming to enhance search efficiency in sequence space. Graph-based smoothing techniques have been introduced to incorporate protein similarity graphs, promoting structural continuity in the fitness landscape \cite{kirjner2023improving, tran2024groot}. These methods aim to reduce abrupt fitness changes caused by minor sequence variations, improving the robustness of optimization algorithms. In general, while experimental and computational strategies each offer advantages and limitations, ongoing research continues to explore ways to enhance protein optimization efficiency while ensuring the reliability and interpretability of computational predictions.

\subsubsection{Ligand-Binding Protein Design}

The design of ligand-binding proteins involves engineering proteins that interact with small molecules such as drugs, metabolites, or signaling compounds. These interactions are central to biological processes, making the design of ligand-binding proteins essential for applications in biosensors, precision medicine, and enzyme-mediated synthesis. Achieving high specificity and affinity for target ligands requires a detailed understanding of the binding site geometry, molecular interactions, and binding thermodynamics, which computational methods have been increasingly employed to model and optimize.

Several benchmark datasets support the evaluation of ligand-binding protein design strategies by providing experimentally validated protein-ligand complexes. PDBbind \cite{PDBbind} offers a curated collection of protein-ligand structures with binding affinity measurements, serving as a reference for affinity prediction and binding site modeling. CSAR (Community Structure Activity Resource) \cite{dunbar2013csar} provides high-resolution crystallographic data with a diverse range of ligand complexes, offering a structured framework for evaluating the specificity of the binding across various protein targets. These datasets are widely used to assess structure-based and sequence-based ligand-binding design approaches.

Computational strategies for the design of ligand-binding proteins typically leverage structural insights from native protein-ligand complexes to refine side chain interactions and backbone conformations, with the aim of improving binding affinity \cite{polizzi2020defined, stark2023harmonic, watson2023novo, dauparas2023atomic}. Many approaches follow a structure-first paradigm, in which protein backbones are generated first, followed by inverse folding techniques \cite{hsu2022learning, gao2023pifold} to identify sequences that can adopt predefined structural conformations. This prioritization of structure prediction before sequence determination allows the design of proteins that conform to specific ligand-binding geometries and can also aid in addressing cases where binding sites are not well defined.

Despite progress, challenges remain, particularly in the structural validation of designed binding modes. Although computational coupling has been applied to generate novel binders by modifying scaffolds and loop geometries, the accuracy and stability of the predicted interactions often require high-resolution experimental confirmation. Furthermore, many design strategies emphasize a limited set of hotspot residues for scaffold placement, which can restrict the exploration of various interaction modes, particularly for targets lacking well-defined pockets or binding clefts \cite{Pocketminer, Gagliardi2023-do}. To address these limitations, alternative approaches have investigated the design of ligand-binding proteins based only on sequence information, bypassing the reliance on structural data \cite{nguyen2024proteinrediff}. These models leverage Protein Language Models (PLMs) to extract functional sequence motifs associated with ligand binding, allowing the design of proteins with desired binding properties even when structural information is unavailable or difficult to obtain.

\subsubsection{Antibody Design}

Antibodies are specialized immune proteins that recognize and bind to specific antigens, such as pathogens, toxins, or diseased cells, marking them for neutralization or destruction. As central components of adaptive immunity, antibodies exhibit high specificity and strong binding affinity, allowing precise immune responses. Their function is primarily dictated by variable regions, particularly Complementarity Determining Regions (CDRs), which mediate antigen recognition. Antibody design focuses on the engineering of antibodies with improved affinity, specificity, and stability, with broad therapeutic applications in cancer, autoimmune disorders, and infectious diseases.

Benchmark datasets play a critical role in evaluating sequence-based and structure-based antibody design. OAS (Observed Antibody Space) \cite{olsen2022observed} provides more than 2 billion immune repertoire sequences from various immune states, species, and individuals, supporting sequence-based antibody modeling. SAbDab (Structural Antibody Database) \cite{dunbar2014sabdab} compiles annotated antibody structures from the PDB, including affinity data and sequence annotations, serving as a key resource for structure-based antibody modeling. AbBind \cite{sirin2016ab} includes 1,101 antibody-antigen mutants in 32 complexes with experimentally measured binding free energy changes, allowing for the study of affinity variations upon mutation. These datasets support the evaluation and development of antibody design approaches, ensuring standardized model assessment across different strategies.

Computational strategies for structure-based antibody design focus primarily on CDRH3 loops, which are highly variable and critical for antigen binding. Two main methodologies have emerged: (i) 3D backbone generation, which designs structurally realistic CDRH3 loops for stable antigen recognition \cite{eguchi2022ig}, and (ii) prediction of binding affinity, which models how sequence modifications impact structural stability and antigen binding energy \cite{shan2022deep}. While structure-based approaches have been widely used, a major challenge is the limited availability of high-resolution antibody structures, which constrains direct structural modeling. To address this, sequence-based models leverage Protein Language Models (PLMs) trained on large-scale immune repertoires to learn patterns associated with high-affinity antibodies. These methods enable de novo antibody generation and affinity maturation without requiring structural information \cite{cohen2023epitopespecific}. Furthermore, sequence-based antibody design can be framed as an optimization problem, where binding affinity serves as the fitness metric, guiding the search for improved variants, similar to optimization strategies used for individual protein properties \cite{wu2025a, lin2025dyab}. A promising direction in antibody design is the integration of sequence and structural information, leading to sequence-structure co-design approaches \cite{jin2022iterative, luo2022antigen, morcillo2023guiding, martinkus2023abdiffuser, zhou2024antigen}. By combining both modalities, these models aim to capture the relationships between sequence variability and structural adaptation, facilitating the precise engineering of antibody specificity, stability, and affinity. This hybrid framework expands the design space beyond naturally occurring immune repertoires, potentially accelerating antibody discovery and improving antigen recognition strategies.

\subsubsection{Enzyme Design}

Enzymes are biological catalysts that accelerate chemical reactions by reducing the activation energy and play a fundamental role in metabolism, signal transduction, and biomolecule synthesis and degradation. Their high specificity and efficiency make them indispensable in both cellular processes and industrial applications, including pharmaceutical production, biofuel synthesis, food processing, and environmental remediation. Enzyme design focuses on the engineering or optimization of enzymes to enhance catalytic efficiency, stability, and substrate specificity for targeted reactions. Traditional approaches such as directed evolution and rational design have been widely applied, but often require extensive experimental screening \cite{doi:10.1021/acscentsci.3c01275}. Computational strategies now enable in silico enzyme prediction and optimization, facilitating a more efficient design process.

Benchmark datasets play a critical role in assessing enzyme design models by providing standardized evaluation frameworks for sequence-function relationships and catalytic performance. UniProt \cite{boutet2007uniprotkb} offers a comprehensive database of protein sequences and functional annotations in various organisms, supporting the design and optimization of sequence-based enzymes. BRENDA \cite{chang2021brenda} compiles detailed enzyme data, including sequence, structure, and kinetic parameters, enabling validation of catalytic efficiency, stability, and substrate specificity in designed enzymes. These benchmarks provide essential resources for the evaluation and development of computational enzyme engineering strategies.

Recent advancements in enzyme design for specific Enzyme Commission (EC) classes have demonstrated promising results, with models generating sequences that closely resemble reference enzymes while maintaining desired catalytic functions \cite{munsamy2022zymctrl, yang2024conditional}. Beyond EC-based classification, alternative approaches have explored de novo enzyme generation by assembling active site and scaffold libraries, followed by refinement algorithms to improve functionality \cite{hossack2023building}. While EC classification provides a structured framework, relying solely on predefined categories may limit model generalization to novel, unseen reactions. To address these challenges, recent research has shifted towards reaction-conditioned enzyme design, which directly models enzyme-substrate relationships rather than relying on predefined EC classifications \cite{mikhael2024clipzyme, hua2025reactzyme}. This approach enables greater flexibility in enzyme generation, allowing models to learn catalytic patterns from reaction-specific data. 

Another key challenge in enzyme sequence design is the limited understanding of enzyme-substrate catalytic mechanisms. Even when designed enzyme sequences fold correctly into 3D structures, catalytic pocket formation and binding interactions often remain poorly characterized. Recent efforts have integrated generative models for enzyme scaffolds, active sites, and protein language models to improve the modeling of enzyme-substrate interaction \cite{hua2024reaction}. By generating enzyme-substrate binding structures, these methods aim to refine the prediction of the catalytic mechanism and support the design of enzymes capable of catalyzing novel reactions.

\subsection{Drug Discovery}

Drug discovery utilizes protein structures and sequences to develop or repurpose small molecules that bind with high specificity and affinity. By analyzing the spatial, physicochemical, and sequence properties of protein targets, researchers aim to design drugs that minimize off-target effects and enhance therapeutic efficacy. Drug discovery is particularly critical for diseases with well-characterized protein targets, including cancer, infectious diseases, and neurodegenerative disorders. This section covers two key tasks: drug design, which focuses on generating novel therapeutic compounds, and drug repurposing, which identifies new applications for existing drugs.

\subsubsection{Drug Design}

Drug design focuses on generating novel small molecules tailored to bind specific protein targets with optimized affinity, stability, and pharmacokinetics. By leveraging both structural and sequence-based information, drug design aims to maximize specificity while minimizing off-target effects, making it essential for treating diseases where precise molecular interactions are crucial, such as cancer and neurodegenerative disorders.

To support model development and evaluation, benchmark datasets provide standardized training and testing resources. CrossDocked \cite{francoeur2020three} consists of cross-docked protein-ligand pairs, challenging models to generalize across diverse binding poses, a crucial ability for designing molecules across multiple protein targets. Binding MOAD (Mother of All Databases) \cite{benson2007binding} offers a large collection of high-resolution protein-ligand complexes with experimentally measured binding affinities, serving as a foundation for refining docking precision and affinity predictions. Together, these datasets facilitate robust evaluation of structure-based drug design (SBDD) models.

Recent advances in PRL have led to deep learning models that generate molecular structures conditioned on 3D representations of protein binding pockets. These models capture the geometric and physicochemical attributes of the binding sites, facilitating the design of small molecules with high affinity and specificity. A predominant strategy in structure-based drug design (SBDD), this pocket-conditioned molecular generation paradigm includes notable models such as Pocket2Mol \cite{peng2022pocket2mol}, ResGen \cite{zhang2023resgen}, PocketFlow \cite{jiang2024pocketflow}, DeepICL \cite{zhung20243d}, and DiffSBDD \cite{schneuing2024structure}, all of which utilize 3D structural data to guide molecular design. By anchoring molecular generation to the spatial and chemical features of protein pockets, these approaches improve ligand design precision and streamline SBDD workflows.

Despite these advances, accurately defining binding pockets for novel protein targets remains a challenge, particularly when structural annotations are unavailable or ambiguous. While many models rely on predefined pockets, emerging efforts are shifting toward whole-protein drug design, allowing models to identify potential binding regions in a data-driven manner without requiring prior annotations \cite{ngo2024multimodal}. This approach broadens the applicability of SBDD to less-characterized targets. Furthermore, a complementary research direction focuses on sequence-based drug design, where molecular structures are generated directly from protein sequences, bypassing the need for 3D structural data \cite{chen2023sequence, creanza2025transformer}. These methods leverage Protein Language Models (PLMs) to extract sequence-level functional insights, offering an alternative strategy for designing drugs against targets with limited or unreliable structural information.

\subsubsection{Drug Repurposing}

Drug repurposing identifies new therapeutic applications for existing drugs, offering a faster and more cost-effective alternative to de novo drug development. This approach is particularly valuable for urgent medical needs, such as pandemics, where traditional drug discovery timelines may be impractical. Broadly, drug repurposing can be categorized into two primary strategies: (i) Virtual screening, which identifies existing drugs with high binding affinity to specific protein targets; and (ii) Drug-Target Interaction (DTI) prediction via link prediction, which models relational networks to uncover novel drug-protein associations.

To facilitate standardized evaluation, benchmark datasets provide essential validation resources for both virtual screening and DTI-based approaches. For virtual screening, DUD-E (Directory of Useful Decoys: Enhanced) \cite{mysinger2012directory} and LIT-PCBA \cite{tran2020lit} are widely used. DUD-E pairs protein targets with both active compounds and decoys, evaluating a model's ability to differentiate true binders from non-binders in complex screening tasks. LIT-PCBA, derived from PubChem bioassays, offers experimentally validated active and inactive compounds, reflecting the challenges encountered in real-world drug repurposing efforts. For DTI-based link prediction, datasets such as PrimeKG \cite{chandak2023building}, Therapeutics Data Commons (TDC) \cite{huang2021therapeutics}, BindingDB \cite{zagidullin2019drugcomb}, and BioSNAP \cite{biosnapnets} provide extensive collections of validated drug-target interactions. These resources support the development and benchmarking of computational models that predict novel drug-target relationships, enabling systematic drug repurposing.

Virtual screening can be further categorized into docking-based and similarity-based methods. Docking-based approaches use molecular docking simulations, such as AutoDock Vina \cite{trott2010autodock}, to computationally dock large drug libraries against protein targets and predict binding affinities. While effective, traditional docking is computationally intensive and limits scalability for high-throughput screening. To overcome these limitations, deep learning-based docking models have been developed, including P2Rank \cite{krivak2018p2rank}, EquiBind \cite{stark2022equibind}, TANKBind \cite{lu2022tankbind}, DiffDock \cite{corso2023diffdock} and HelixDock \cite{liu2023pre}, to predict binding poses and affinities with reduced computational costs. These models leverage 3D protein-ligand complexes to learn structural binding interactions and improve docking accuracy. In contrast, similarity-based approaches rely on the ``similarity principle'', which assumes that structurally similar molecules exhibit similar biological properties \cite{bender2009similar}. Rather than explicitly docking compounds, these methods use learned molecular representations to rapidly identify structurally related drug candidates. Recent advances in PRL have enabled drug and protein embeddings to be mapped into a shared representation space, facilitating efficient similarity searches. Unlike docking-based methods that depend on 3D structural data, similarity-based approaches do not require explicit protein-ligand modeling. Instead, they employ pre-trained models, such as Protein Language Models (PLMs) and Molecular Graph Neural Networks (GNNs), to encode proteins and ligands separately. Notable examples include ConPLex \cite{singh2023contrastive}, DrugCLIP \cite{gao2023drugclip}, and SPRINT \cite{mcnutt2024sprint}, which use contrastive learning to map proteins and drugs into a joint representation space, allowing for rapid and scalable virtual screening without the computational cost of docking simulations.

Beyond virtual screening, DTI-based link prediction offers an alternative paradigm for drug repurposing by formulating drug-target interactions as a graph-based link prediction task. In this approach, drugs and proteins are represented as nodes, while known interactions form edges. A key challenge in DTI-based models is how the drug and protein nodes are represented, as the quality of the node embeddings directly impacts model performance. Recent methods have introduced various node embedding strategies, which can be classified into sequence-based and multimodal embeddings. Sequence-based embeddings use PLMs for protein sequences and SMILES-based transformers for molecular representations \cite{daza2023bioblp, djeddi2023advancing}, encoding drugs and proteins based on their primary sequences without requiring structural data. Meanwhile, multimodal embeddings integrate diverse sources of biological information, combining sequence, structure, and auxiliary biological data, such as functional annotations, side-effect similarities, and pathway interactions, to construct more comprehensive representations \cite{dang2025multimodal, dong2023multi}. By capturing both structural and functional relationships, these models improve the predictive accuracy of link prediction tasks, making DTI-based drug repurposing a valuable complement to virtual screening approaches by providing insights into drug-target associations that may not be captured through traditional binding-based models.
\section{Future Directions and Open Challenges}
\label{sec:future}
\begin{figure*}[ht]
    \centering
    \includegraphics[width=\textwidth]{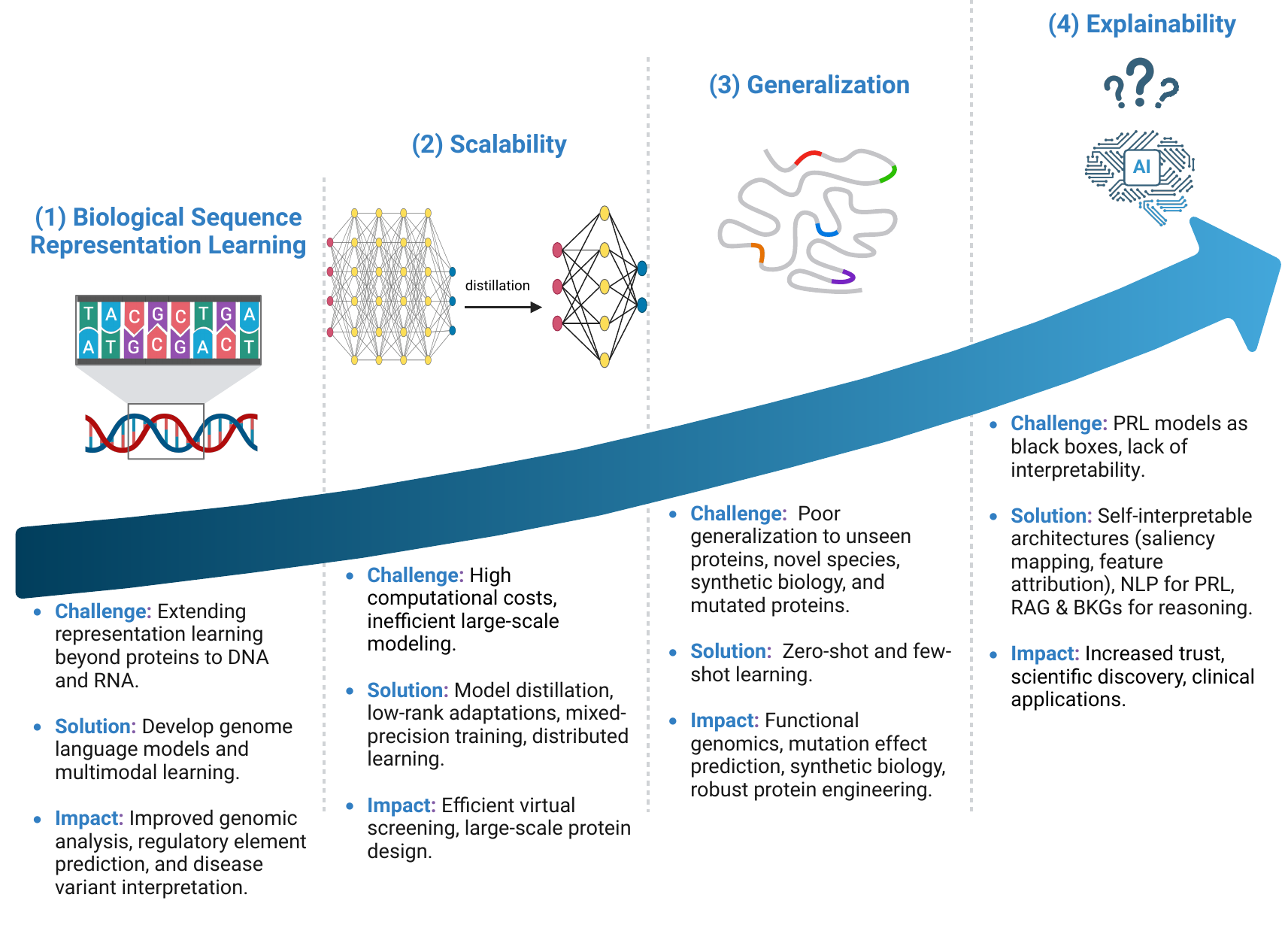}
    \caption{\label{fig:future_directions} Key future directions in Protein Representation Learning (PRL). The figure outlines four critical challenges and potential advancements: (i) Expanding PRL to DNA and RNA representation learning to leverage shared methodologies while addressing unique challenges, (ii) Enhancing scalability to improve computational efficiency and accessibility of large-scale models, (iii) Strengthening generalization to ensure robustness across unseen proteins and genetic variations, and (iv) Advancing explainability to improve model interpretability and facilitate trust in biological and biomedical applications.}
\end{figure*}

While PRL has made significant strides, several challenges and opportunities remain that could shape the future development of the field. This section explores four key directions that we believe will drive further advancements: extending PRL methodologies to other biological sequences such as DNA and RNA to uncover regulatory mechanisms and gene functions, improving model explainability to enhance interpretability and trust, scaling PRL models to improve efficiency and accessibility, and enhancing generalization to improve robustness across unseen and mutated proteins, as summarized in \cref{fig:future_directions}.

\subsection{Biological Sequence Representation Learning}

After significant advancements in Protein Representation Learning (PRL), a natural progression is the extension of these methods to other biological macromolecules, such as DNA and RNA. Learning meaningful representations of DNA and RNA sequences is critical for understanding gene regulation, epigenetics, alternative splicing, and disease mechanisms. By applying methodologies developed in PRL to these biological sequences, researchers can leverage the power of deep learning models to extract insights that may otherwise remain hidden within complex genomic and transcriptomic data.

The representation learning of DNA and RNA shares foundational similarities with the protein representation learning, as both involve extracting meaningful features from biological sequences to predict function, interactions, and structural properties. Like proteins, DNA and RNA sequences follow a linear arrangement of biological ``letters'' that encode essential biological information. Advances in self-supervised learning and transformer-based architectures, which have successfully modeled protein sequences, can be applied similarly to DNA and RNA to capture sequence patterns, evolutionary conservation, and functional motifs. Recent efforts to develop genome language models have shown promising results, demonstrating their potential in learning meaningful representations of genomic sequences and improving the prediction of gene function, regulatory elements, and chromatin accessibility \cite{nguyen2024sequence, brixi2025genome, Consens2025}.

However, DNA and RNA present unique challenges distinct from proteins. Unlike proteins, which have a well-defined alphabet of 20 amino acids and a structured hierarchical organization, DNA and RNA sequences consist of only four nucleotides (A, T/U, G, and C) but often span much longer sequences, reaching tens of thousands or even millions of base pairs. This length disparity makes learning effective representations computationally intensive and necessitates efficient sequence-compression strategies. Furthermore, while proteins fold into stable three-dimensional structures that largely dictate function, DNA and RNA often exhibit dynamic secondary and tertiary structures influenced by sequence context, environmental factors, and cellular conditions. Capturing these structural variations requires novel modeling approaches that integrate both sequence-based and structural information. Furthermore, the function of DNA and RNA is strongly modulated by epigenetic modifications, chromatin accessibility, and interactions with regulatory proteins. Unlike proteins, where function is often inferred from sequence and structure, the functional state of a genomic region is highly context-dependent, requiring models to account for regulatory mechanisms beyond the primary sequence.

\subsection{Scaling PRL Models for Large-Scale Applications}
As biological datasets continue to grow in size and complexity, the scalability of PRL models remains a significant challenge. Many existing methods struggle with large-scale protein representations, requiring substantial computational resources that limit accessibility and real-world applicability. Addressing these challenges requires advancements in model efficiency and optimization techniques.

One promising direction for improving scalability is model distillation, a technique that has seen significant progress in Natural Language Processing (NLP) and Computer Vision (CV). In NLP, approaches such as DistilBERT \cite{sanh2019distilbert} and TinyBERT \cite{jiao2019tinybert} have successfully compressed large transformer models while preserving much of their predictive power. Similarly, in CV, TinyViT \cite{wu2022tinyvit} and DearKD \cite{chen2022dearkd} leverage distillation to create more compact and computationally efficient models. These techniques demonstrate the potential of distilling knowledge from large, computationally expensive models into smaller, more efficient variants without a significant loss of accuracy. Recent efforts have explored applying model distillation techniques to Protein Representation Learning (PRL) to improve scalability and efficiency. For instance, OpenFold \cite{ahdritz2024openfold} employs knowledge distillation to create a more computationally efficient version of AlphaFold.

Applying model distillation to PRL could enhance efficiency by enabling smaller models to retain essential biological insights while significantly reducing memory and computational overhead. This would be particularly beneficial for large-scale virtual screening, protein design, and low-data scenarios where fine-tuning massive models is impractical. Furthermore, integrating distillation with techniques such as low-rank adaptations, mixed-precision training, and distributed learning could further enhance efficiency while maintaining high performance.

\subsection{Improving Generalization of PRL Models}

While scalability focuses on computational efficiency, generalization remains a critical challenge in PRL. Many models struggle to generalize to unseen proteins, such as those from novel species, synthetic biology applications, or de novo protein designs, limiting their applicability in biomedical and biotechnological research. Beyond handling completely novel proteins, PRL models must also account for mutated proteins, where small substitutions, deletions, or insertions of amino acids can significantly alter the function, stability, or interactions of a protein. Understanding how mutations affect learned representations is crucial for applications such as disease mutation analysis, where genetic variants influence protein function; protein engineering, where specific mutations can optimize protein stability and activity; and drug resistance prediction, where mutations in pathogens impact therapeutic efficacy. However, the availability of labeled data for both novel proteins and mutant proteins is often extremely limited, making it difficult to train models that can reliably predict their properties.

To address this, zero-shot and few-shot learning approaches need to be developed, enabling PRL models to generalize effectively to unseen proteins and mutations with minimal supervision. Recent studies have demonstrated promising results in this area, showing that pretrained PRL models can infer functional properties and structural effects of mutations even in data-scarce settings, improving predictive performance in mutation effect prediction and de novo protein engineering \cite{tan2024retrieval, cheng2024zero}.

\subsection{Enhancing Explainability in PRL Models}
Despite their success, many state-of-the-art PRL models operate as black boxes, making it difficult to interpret how they learn protein representations and make predictions. This lack of transparency hinders scientific discovery, limits trust in model outputs, and reduces their applicability in critical fields such as medicine, biotechnology, and drug development. As PRL models become increasingly complex, using deep neural networks, large-scale pre-training, and self-supervised learning, there is a growing need for approaches that can make their predictions more interpretable and biologically meaningful.

An approach to enhancing explainability in PRL is the development of self-interpretable architectures that inherently capture biologically meaningful representations. Explainable AI techniques, such as attention mechanisms, saliency mapping, and feature attribution, allow models to highlight critical sequence motifs, structural domains, and evolutionary patterns that drive their predictions. By identifying key residues, structural elements, and physicochemical properties relevant to specific functions or interactions, these methods improve model transparency and reliability. This enhanced interpretability facilitates applications in functional protein annotation, biomarker discovery, and rational protein design. Recent studies have explored these approaches, applying various Explainable AI techniques to improve the interpretability of PRL models \cite{gong2025multigrandti, medina2024interpretable}.

Another promising direction is to leverage advances in Natural Language Processing (NLP) and Large Language Models (LLM) to make PRL models more accessible and interpretable. Inspired by recent successes in LLM, PRL models could be designed to generate natural language descriptions of protein functions, structural properties, and interactions, making predictions more intuitive for researchers in biology, medicine, and drug development. Beyond simple text-based outputs, reasoning-based approaches could further enhance explainability by allowing models to logically infer protein functions and molecular interactions rather than relying solely on statistical pattern recognition. Techniques such as retrieval-augmented generation (RAG) and biomedical knowledge graphs (BKGs) could serve as foundational tools for reasoning, enabling models to retrieve relevant biological context from structured databases and scientific literature. Instead of merely generating descriptions, PRL models could use retrieved biological relationships and prior knowledge to construct logical justifications for their predictions, improving their transparency and reliability. Recent studies have already begun exploring these approaches, demonstrating promising results in improving the interpretability of the model and providing biologically meaningful insights \cite{ma2025prottexstructureincontextreasoningediting, wang2025prot2chatproteinllmearly}. By integrating LLM-based reasoning, knowledge retrieval, and structured biological insights, PRL models could evolve into more explainable, evidence-driven systems, facilitating their adoption in research and clinical applications.

\section{Conclusion}
\label{sec:conclusion}

Protein representation learning (PRL) has emerged as a transformative approach to advance our understanding of proteins and to address critical challenges in molecular biology, biotechnology, and medicine. This review provides a comprehensive overview of PRL methodologies, categorized into five main approaches: feature-based, sequence-based, structure-based, multimodal, and complex-based. By analyzing the strengths and limitations of each approach, we offer insights to guide the selection of appropriate PRL strategies based on specific biological and computational requirements, ensuring their effective application across diverse research domains.

Beyond methodological advancements, we highlight key databases that underpin PRL research and discuss its diverse applications, including protein property prediction, structure modeling, protein design, and drug discovery. Despite substantial progress in PRL, several open challenges remain, which present opportunities for future research. Key directions for advancement include extending PRL methodologies to genomic representation learning, enhancing model scalability and generalization, and improving interpretability to foster broader accessibility and adoption. Addressing these challenges will be crucial to refining PRL methodologies and expanding their impact on biological and biomedical research.

\bibliography{main}
\bibliographystyle{plain}

\end{document}